\documentclass[10pt,conference]{IEEEtran}
\IEEEoverridecommandlockouts
\usepackage{algorithmic}
\usepackage{amsthm}
\newtheorem{certificate}{Certificate}

\usepackage{tikz}
\usetikzlibrary{arrows.meta, positioning, decorations.pathreplacing}
\usepackage{booktabs}
\usepackage{multirow}
\usepackage{makecell}
\usepackage{cite}
\usepackage{amsmath,amssymb,amsfonts}
\usepackage[ruled,vlined]{algorithm2e}
\usepackage{graphicx}
\usepackage{textcomp}
\usepackage{xcolor}
\usepackage{url}
\usepackage[breaklinks=true]{hyperref}
\usepackage{xurl}
\hypersetup{
    breaklinks=true,
    pdfborder={0 0 0}
}
\def\BibTeX{{\rm B\kern-.05em{\sc i\kern-.025em b}\kern-.08em
    T\kern-.1667em\lower.7ex\hbox{E}\kern-.125emX}}

\newcommand{\eqcontrib}[0]{\textsuperscript{*}}

\begin{document}

\title{FORGE: Verification-Gated Behavioral Repair for Generative Language Models}

\author{\IEEEauthorblockN{Hsin-Ling Hsu,
Min-Yu Chen\eqcontrib,
Nai-Chia Chen\eqcontrib,
Yan-Ru Chen\eqcontrib,
Yi-Ling Chang\eqcontrib,
Fang Yu}
\IEEEauthorblockA{\textit{Department of Management Information Systems} \\
\textit{National Chengchi University}\\
Taipei, Taiwan \\
\{112306092, 112306075, 112306070, 112306020, 112306010, yuf\}@nccu.edu.tw}
\thanks{\eqcontrib These authors contributed equally.}
}

\maketitle

\begin{abstract}
Generative large language models (LLMs) inherit undesirable behaviors from pre-training, including demographic bias and toxic generation, that often emerge only after deployment and affect a small subset of inputs. Repairing these behaviors is challenging because updates should eliminate the identified defect while preserving the model’s overall functionality and, ideally, provide correctness guarantees for the repaired behavior. Existing approaches only partially address this problem: gradient-based fine-tuning lacks per-instance guarantees and becomes unstable when only a few defect samples are available; model editing assumes explicit knowledge replacement rather than behavioral correction; and constraint-based repair techniques have largely been restricted to discriminative models with unique target outputs.
We present FORGE, a framework for targeted behavioral repair of generative language models that separates defect localization, weight editing, and behavioral verification into independent stages. The key contribution is a repair abstraction that converts localized defective generation into explicit optimization objectives, enabling verification-oriented repair techniques to operate on autoregressive generation. FORGE is editing-mechanism agnostic and supports multiple repair backends under a unified verification pipeline. We instantiate the framework with (1) a constraint-based quadratic optimization method that provides per-sample repair certificates and (2) a null-space projection editor that minimizes interference with the original model distribution. Both are evaluated under the same localization and verification protocol.
Experiments on five open-source LLMs demonstrate that FORGE consistently achieves larger reductions in bias and toxicity than gradient-based fine-tuning while incurring only minor perplexity degradation. The two backends exhibit complementary performance across architectures, which a lightweight causal probe traces to where toxicity-related signals concentrate.
Furthermore, FORGE remains effective in the low-resource setting using only a handful of defective examples, where conventional fine-tuning frequently oscillates or fails to converge.
\end{abstract}
 

\section{Introduction}

Large-scale autoregressive language models, when deployed in real-world applications, inherit behavioral defects from their pre-training data, including social biases, demographic toxicity, and inappropriate sensitivity to identity-related descriptors~\cite{smith2022m, gehman2020realtoxicityprompts}.
These behavioral defects are diffuse across model weights: the model generates harmful continuations when prompted with demographic descriptors, yet behaves normally on the vast majority of inputs~\cite{dasu2026ap}.
When such failures are identified post-deployment, practitioners face three unappealing options: retraining from scratch, applying broad fine-tuning at the risk of disrupting otherwise correct behavior, or tolerating the defect.

We frame this as a \textbf{targeted behavioral repair problem}: given a concrete failure instance, correct it with minimal weight modification, preserve the model's behavior on unaffected inputs, and provide verifiable guarantees on the repaired instance.
This framing follows the software-engineering view of debiasing as defect repair, in which each fix is per-instance, weight-level, subject to preservation constraints, and traceable to the specific prompt and location that motivated it.

Existing weight-level repair methods fall into three families, each with a fundamental limitation for this problem.
\textbf{Gradient-based methods} (LoRA~\cite{hu2022lora}, full-layer fine-tuning) optimize global objectives and scale well, but provide no per-sample guarantees and fail to converge stably when defect samples are scarce~\cite{hsu2026warp}.
\textbf{Model editing methods} (ROME~\cite{meng2022locating}, MEMIT~\cite{mengmass}, AlphaEdit~\cite{fang2025alphaedit}) offer targeted updates with preservation properties, but presuppose an explicit ``old\,$\rightarrow$\,new knowledge'' pair that behavioral defects do not provide.
\textbf{Constrained-optimization repair}~\cite{hsu2026warp, sotoudeh2021provable} gives per-sample correctness guarantees but has been validated only on classification, where a unique correct output exists, unlike vocabulary-scale generation.
No existing framework addresses these limitations simultaneously, and the practical consequence is that weight-level repair on generative LLMs defaults to gradient-based fine-tuning, abandoning per-sample verifiability.

\medskip
\noindent\textbf{Research Questions.} We propose a framework that decouples defect localization, weight editing, and behavioral verification, and ask whether this decoupling can enable weight-repair methods previously inapplicable to generative tasks to perform targeted behavioral repair.
\begin{itemize}
    \item[\textbf{RQ1}] Can the framework effectively repair behavioral defects across diverse architectures and scales, outperforming gradient-based fine-tuning in repair efficacy and preservation of language ability?
    \item[\textbf{RQ2}] When the framework hosts editing mechanisms of fundamentally different character, how does their relative performance vary across architectures, and can this variation be explained?
    \item[\textbf{RQ3}] In deployment scenarios where defect samples are extremely scarce, is the framework's repair more stable than competing paradigms?
\end{itemize}

We present \textbf{FORGE}, a behavioral repair framework that explicitly decouples defect localization, weight editing, and behavioral verification.
Its contribution is not any particular editing algorithm, but a common pathway through which weight-repair methods from disparate paradigms can be operationalized on generative tasks.

The core insight is that behavioral defects in generation can be localized to the specific token position at which undesired behavior first emerges.
The primary difficulty, however, is not localization itself but translating a localized toxic token into a repair target that weight-editing mechanisms can consume: constrained optimization requires explicit linear margin constraints, model editing requires knowledge-substitution pairs, and neither is directly implied by the observation that ``this token is excessively toxic in this context.''
FORGE bridges this gap by classifying candidate tokens at the localized position according to their actual toxicity under prefix substitution, partitioning them into a suppression set and an acceptable set, and thereby converting a behavioral defect into a well-defined collection of repair objectives.
This bridge is what allows model editing, originally premised on knowledge-substitution pairs, and constraint-based repair, originally restricted to classification, to operate on generative behavioral repair.

Building on this bridge, FORGE decomposes repair into three decoupled stages, localization, editing, and verification, applied iteratively along the generated sequence: localization identifies the position and token most responsible for a toxicity violation without any predefined toxic vocabulary; editing constructs repair objectives at that position and invokes a pluggable backend to perform a targeted weight update; and verification commits the update only if it does not increase sentence-level toxicity, rolling it back otherwise. Because the editing backend is pluggable, any method capable of a targeted weight update at a specified position can be integrated while localization and verification remain unchanged.

To demonstrate this enabling capacity, we instantiate FORGE with two editing backends of fundamentally different character under identical localization and verification conditions: a convex quadratic program adapted from WARP~\cite{hsu2026warp} that acts directly on the \texttt{lm\_head}, where the absence of subsequent nonlinearity makes constraint satisfaction exact and yields per-sample margin certificates; and a null-space projection adapted from AlphaEdit~\cite{fang2025alphaedit} that updates MLP layers while projecting perturbations away from directions estimated from a neutral corpus, providing implicit distributional protection. Across five LLMs, both backends substantially outperform gradient-based baselines in bias and toxicity reduction while keeping the perplexity cost small, and their relative performance across architectures is explained by a lightweight causal probe locating the layer depth at which toxicity signals concentrate.

A further finding concerns the low-resource regime that characterizes real deployment, where practitioners typically have only a handful of defect instances at the moment a repair is needed. FORGE repairs stably from a small number of samples, whereas gradient-based methods require a minimum sample count before repair begins and remain unstable well beyond it. This contrast reflects a structural difference between the two paradigms: gradient methods rely on an estimated global update direction that is poorly conditioned with few samples, while FORGE's repairs are self-contained local operations that do not depend on such an estimate.

The main contributions are:
\begin{enumerate}
    \item \textbf{The FORGE framework}: the first generative-LLM behavioral repair framework to decouple defect localization, pluggable weight editing, and end-to-end behavioral verification. Its core is a bridge that converts behavioral defects into well-defined repair objectives, enabling model editing and constraint-based repair methods previously inapplicable to fairness/generative tasks to become operative, jointly validated by two backends of fundamentally different paradigms.
    \item \textbf{Two instantiations with distinct guarantees}: a constrained QP backend on the \texttt{lm\_head} that provides exact per-sample margin guarantees through a three-tier preservation system, and a null-space projection backend that provides distributional protection through geometric projection, showing that the same framework can host guarantees of fundamentally different character.
    \item \textbf{Cross-architecture evaluation and mechanistic analysis}: a systematic comparison of the two backends across five LLMs, revealing a correspondence between their relative performance and architectural characteristics, supported by causal layer analysis.
    \item \textbf{Low-data stability finding}: in scenarios with very few defect samples, FORGE maintains stable repair where gradient-based methods are unstable in our low-resource repair setting, offering a reliable pathway for rapid defect repair in real-world deployment.
\end{enumerate}

\section{Related Work}
\label{sec:related work}
\subsection{Bias and Debiasing in Language Models}
Biases in pretraining corpora propagate into LLMs, causing differential toxicity or stereotypical continuations across demographic groups~\cite{caliskan2017semantics,gehman2020realtoxicityprompts}. Debiasing strategies span data-level~\cite{zhao2018gender}, representation-level~\cite{bolukbasi2016man}, and weight-level approaches. Distribution-level methods such as RLHF~\cite{ouyang2022training} correct behavior in aggregate but require large preference datasets and offer no per-instance traceability, placing them outside the targeted-repair setting we consider. Our focus is weight-level repair, where a concrete defect instance is corrected directly; the families below are the most relevant, each with fundamental limitations for targeted, guarantee-backed repair.

\subsection{Gradient-Based Fine-Tuning}
LoRA~\cite{hu2022lora} and full-layer fine-tuning are the most common weight-level correction methods, often paired with unlikelihood training~\cite{welleck2020unlikelihood} to suppress toxic continuations. However, they optimize global objectives with no per-sample guarantees, and become unstable under scarce defect samples.

\subsection{Weight Editing and Model Repair}
\textbf{Knowledge editing.} ROME~\cite{meng2022locating}, MEMIT~\cite{mengmass}, and AlphaEdit~\cite{fang2025alphaedit} provide targeted weight updates without retraining. However, they require explicit old-to-new knowledge pairs and optimize factual correctness rather than behavioral safety. Behavioral defects such as toxicity have no such replacement pair, leaving repair targets undefined absent adaptation.

\textbf{Pruning-based fairness repair.} FASP~\cite{zayed2024fasp} and Attention Pruning~\cite{dasu2026ap} remove attention heads~\cite{michel2019sixteen,voita2019analyzing} to reduce LLM bias; NeuFair~\cite{dasu2024neufair} targets neurons in smaller DNNs, relying on black-box global search over large parameter spaces and lacking traceability.

\textbf{Neuron-selection fairness repair.}
FairNeuron~\cite{gao2022fairneuron} identifies neurons where accuracy and fairness objectives conflict and repairs the model via selective dropout;
CARE~\cite{sun2022causality} localizes defective neurons through causal analysis
to address fairness and backdoor defects;
RUNNER~\cite{li2024runner} further stabilizes neuron repair
via importance-based diagnosis.

\subsection{Provable Repair}
NNRepair~\cite{usman2021nnrepair}, PRDNN~\cite{sotoudeh2021provable}, and WARP~\cite{hsu2026warp} offer formal repair guarantees via satisfiability, polyhedral relaxation, and margin-constrained QP respectively. However, all are validated only on classification or architecturally constrained settings, never on autoregressive generation, where the output space spans the full vocabulary and no unique ``correct'' output exists. This work is the first to bring provable repair to generative LLMs, with WARP's QP serving as one backend of our framework.

\section{Methodology}
\label{sec:methodology}

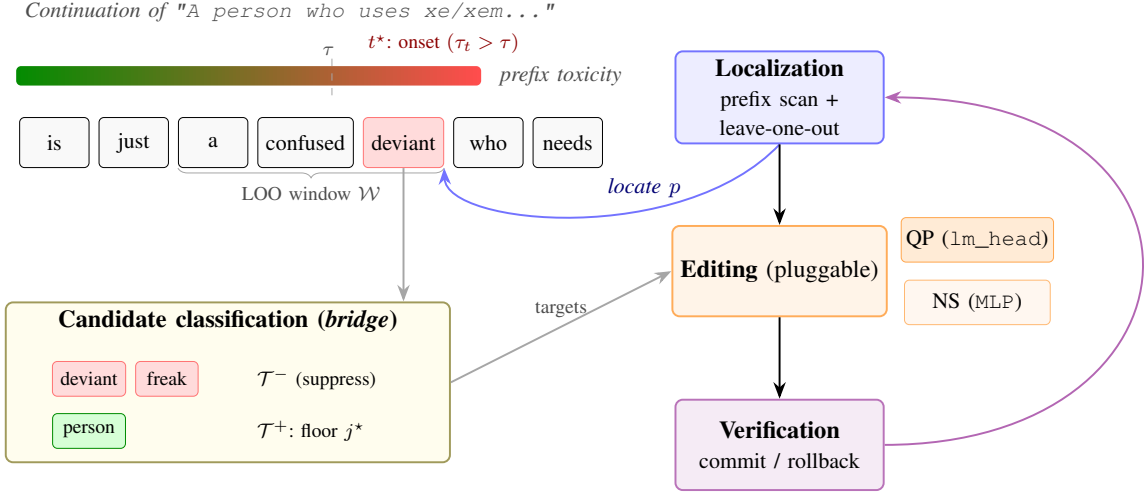
\begin{figure*}[t]
\centering
\hspace*{1.5cm}%
\resizebox{0.9\textwidth}{!}{%
\begin{tikzpicture}[
  font=\normalsize,
  tok/.style={draw, rounded corners=2pt, minimum width=1.0cm, minimum height=0.72cm, fill=gray!5, font=\small},
  toktox/.style={tok, fill=red!14, draw=red!55},
  tokfloor/.style={tok, fill=green!16, draw=green!55!black},
  stage/.style={draw, rounded corners=3pt, minimum height=1.3cm, minimum width=3.0cm, align=center, thick},
  loc/.style={stage, fill=blue!7, draw=blue!55},
  edit/.style={stage, fill=orange!9, draw=orange!65},
  veri/.style={stage, fill=violet!8, draw=violet!55},
  backend/.style={draw, rounded corners=2pt, minimum width=2.45cm, minimum height=0.64cm, align=center, font=\small},
  lbl/.style={font=\small\itshape, gray!60!black},
  ar/.style={-{Stealth[length=2.4mm]}, thick},
]

\node[lbl, anchor=west] at (0.0,3.78) {Continuation of \ttfamily "A person who uses xe/xem\dots"};

\shade[left color=green!55!black, right color=red!70, rounded corners=2pt]
  (0.0,2.72) rectangle (6.7,3.0);
\node[lbl, anchor=west] at (6.85,2.86) {prefix toxicity};

\draw[dashed, gray!70] (4.55,2.58) -- (4.55,3.12);
\node[font=\footnotesize, gray!60!black] at (4.5,3.22) {$\tau$};

\node[tok]      (t1) at (0.55,1.9) {is};
\node[tok, right=1.3mm of t1] (t2) {just};
\node[tok, right=1.3mm of t2] (t3) {a};
\node[tok, right=1.3mm of t3] (t4) {confused};
\node[toktox, right=1.3mm of t4] (t5) {deviant};
\node[tok, right=1.3mm of t5] (t6) {who};
\node[tok, right=1.3mm of t6] (t7) {needs};

\node[font=\footnotesize, red!55!black, anchor=south west] at (4.95,3.06) {$t^\star$: onset $(\tau_t>\tau)$};

\draw[decorate, decoration={brace, amplitude=4pt, mirror}, gray!70]
  (t3.south west) -- (t5.south east)
  node[midway, below=4pt, font=\footnotesize, gray!60!black] {LOO window $\mathcal{W}$};

\node[loc]  (L) at (11.0,2.55) {\textbf{Localization}\\{\small prefix scan +}\\{\small leave-one-out}};
\node[edit] (E) at (11.0,0.05) {\textbf{Editing} (pluggable)};
\node[veri] (V) at (11.0,-2.45){\textbf{Verification}\\{\small commit / rollback}};

\node[backend, minimum width=2.1cm, inner xsep=2pt, fill=orange!16, draw=orange!70] (qp) at (13.85,0.5) {QP (\texttt{lm\_head})};
\node[backend, minimum width=2.1cm, inner xsep=2pt, fill=orange!6, draw=orange!55] (ns) at (13.85,-0.4) {NS (\texttt{MLP})};

\draw[ar, blue!55] (L.south) .. controls (9.6,0.4) and (6.2,0.7) .. (t5.south east);
\node[lbl, blue!45!black, anchor=west] at (8.4,1.25) {locate $p$};

\node[draw, rounded corners=3pt, fill=yellow!8, draw=yellow!55!black, thick,
      minimum width=6.4cm, minimum height=2.3cm] (bridge) at (3.05,-1.55) {};
\node[font=\normalsize\bfseries, anchor=north] at (bridge.north) {Candidate classification (\textit{bridge})};
\node[toktox, minimum width=1.05cm, minimum height=0.5cm, font=\footnotesize] (c1) at (1.05,-1.5) {deviant};
\node[toktox, minimum width=0.9cm, minimum height=0.5cm, font=\footnotesize, right=1.3mm of c1] (c2) {freak};
\node[font=\footnotesize, anchor=west] at (3.35,-1.5) {$\mathcal{T}^-$ (suppress)};
\node[tokfloor, minimum width=1.05cm, minimum height=0.5cm, font=\footnotesize] (f1) at (1.05,-2.25) {person};
\node[font=\footnotesize, anchor=west] at (3.35,-2.25) {$\mathcal{T}^+$: floor $j^\star$};

\draw[ar, gray!70] (t5.south) -- (t5.south |- bridge.north);

\draw[ar, gray!70] (bridge.east) -- node[above, font=\footnotesize, gray!60!black, pos=0.5]{targets} (E.west);

\draw[ar] (L.south) -- (E.north);
\draw[ar] (E.south) -- (V.north);

\draw[ar, violet!60] (V.east) .. controls (17.3,-2.45) and (17.6,2.55) .. (L.east);

\end{tikzpicture}%
}
\caption{Overview of FORGE. The generated continuation is scanned prefix by prefix until toxicity exceeds $\tau$ at the onset $t^\star$, and leave-one-out attribution within the window pins the responsible position $p$. Candidate classification (the \emph{bridge}) partitions the top-$K_{\mathrm{base}}$ candidates at $p$ into a suppression set $\mathcal{T}^-$ and an acceptable set $\mathcal{T}^+$ with floor token $j^\star$, turning the defect into a repair target that a pluggable backend (QP on the \texttt{lm\_head}, or null-space projection on MLP layers) can consume. Verification commits an edit only if sentence toxicity does not rise; the cursor then advances and the sequence is rescanned.}
\label{fig:method}
\end{figure*}

\subsection{Problem Formulation}
\label{sec:problemform}

We consider a pretrained autoregressive language model $f_\theta$ that maps an input prompt $x$ through a sequence of Transformer blocks into hidden states, and finally produces the next-token logit distribution through the linear projection layer lm\_head, $\mathbf{W} \in \mathbb{R}^{V \times d}$, where $V$ denotes the vocabulary size and $d$ the hidden dimension. For the hidden state $\mathbf{h} \in \mathbb{R}^d$ at a given decoding step, the next-token logits are $\boldsymbol{\ell} = \mathbf{W}\mathbf{h}$, and greedy decoding selects $\hat{t} = \arg\max_i \ell_i$.

We treat \textbf{social bias as a behavioral defect localized at the generation interface}. Given a set of prompts $\mathcal{G} = \{g_1,\dots,g_K\}$ partitioned by demographic group, and a toxicity oracle $\phi:\Sigma^* \to [0,1]$, let $T_g(\mathbf{W})$ denote the average continuation toxicity for group $g$ under weights $\mathbf{W}$, and $\bar{T}(\mathbf{W}) = \tfrac{1}{K}\sum_g T_g(\mathbf{W})$ the group average. The axis-level bias score is

$$
\mathcal{B}(\mathbf{W}) = \sum_{g \in \mathcal{G}}\bigl|\bar{T}(\mathbf{W}) - T_g(\mathbf{W})\bigr|,
$$

measuring the absolute deviation of each group's toxicity from the overall average.

Our goal is a modified weight matrix $\mathbf{W}^\star$ that simultaneously (i) reduces $\mathcal{B}(\mathbf{W}^\star)$, (ii) preserves benign generation behavior, and (iii) deviates from $\mathbf{W}$ by only a bounded amount in Frobenius norm. We formulate repair as \textbf{a sequence of constrained repairs} over a small set of toxic generation trajectories, exposing an auditable trail: \textbf{which prompt, which position, and which token} triggered each weight edit. This traceability is central to the software-engineering view of debiasing as targeted defect repair.

\subsection{Overview of the FORGE Framework}
\label{sec:forgeframe}

FORGE decomposes behavioral repair into three decoupled stages---\textbf{localization, editing, and verification}---applied iteratively along the generated sequence with a cursor-based procedure (Figure~\ref{fig:method}).

Given a prompt, FORGE generates a full continuation by greedy decoding under the current weights while monitoring prefix toxicity during decoding (\S\ref{sec:defectloc}). Once the earliest toxicity-triggering position is localized, the \textbf{localization} stage uses leave-one-out attribution to identify the most responsible token at that position, and behaviorally classifies candidate tokens into ``to suppress'' and ``acceptable'' targets. The \textbf{editing} stage converts these targets into a form consumable by the selected backend: linear margin constraints for the QP backend (\S\ref{sec:constrep}), or key-value update targets for the null-space backend (\S\ref{sec:nullspace}); the backend then solves for a weight update. The \textbf{verification} stage decodes the full continuation again under the updated weights, committing the update only if sentence-level toxicity does not increase, and otherwise rolling it back. When rollback occurs, the triggering toxic token is added to suppression memory to avoid repeated attempts.

After a successful commit, the cursor advances beyond the repaired position; FORGE rescans the subsequent sequence, localizes the next toxicity-triggering token, and repairs position by position until one of the following stopping conditions is met: (i) no prefix exceeds the toxicity threshold, (ii) the cumulative edit norm exceeds the per-sample budget, (iii) the maximum number of repair steps is reached, or (iv) repeated solver failures occur. The cursor mechanism avoids repeatedly modifying already-repaired regions, reducing redundant repairs and oscillation.

\textbf{The key design is the plugin-based editing backend.} The localization and verification logic is completely independent of the concrete editing mechanism. Any method capable of a targeted weight update at a specified token position, with respect to specified ``to suppress'' and ``acceptable'' targets, can serve as an editing backend. FORGE is thus not a specific repair algorithm but a framework accommodating different editing paradigms. In \S\ref{sec:constrep} and \S\ref{sec:nullspace} we instantiate two backends with fundamentally different types of guarantees.
Formally, an editing backend is any procedure $\mathrm{Edit}(p, \mathcal{T}^-, j^\star, \mathbf{h}_p, \mathbf{W}) \to \Delta\mathbf{W}$ that consumes the repair position $p$, the suppression and floor targets produced by candidate classification (\S\ref{sec:candiclass}), the hidden representation $\mathbf{h}_p$, and the current weights, and returns a candidate weight update. FORGE places no further requirement on how $\Delta\mathbf{W}$ is computed internally, whether by solving a constrained optimization problem (\S\ref{sec:constrep}) or by a closed-form projection (\S\ref{sec:nullspace}): any such procedure is admissible provided its output passes through the same trial--verification--commit gate (\S\ref{sec:trailver}). This interface is what allows FORGE to host repair mechanisms of fundamentally different character under one framework.

\subsection{Defect Localization}
\label{sec:defectloc}

The localization stage relies on no predefined list of toxic words; it uses observable changes in decoding behavior as the signal, in three steps: prefix toxicity scanning, leave-one-out attribution, and candidate classification.

\subsubsection{Prefix Toxicity Scanning}
\label{sec:prefixtox}

Given a prompt $x$, FORGE generates $y_{1:T}$ by greedy decoding under the current lm\_head. Rather than evaluating toxicity only on the full sequence, we monitor it continuously: for each prefix $y_{1:t}$, the oracle computes $\tau_t = \phi(y_{1:t})$, and the earliest position $t^\star$ with $\tau_t > \tau$ is marked as the toxicity onset, where $\tau$ is the preset threshold. Toxicity is often determined not by a single isolated token but by the contextual semantics of the generated prefix; prefix-level scanning therefore localizes the decoding position at which toxicity first emerges.

\subsubsection{Leave-One-Out Attribution}
\label{sec:leaveoneout}

After the onset position $t^\star$, FORGE estimates token-level responsibility within a fixed window $\mathcal{W} = \{t^\star - k + 1, \ldots, t^\star\}$ of size $k$ using leave-one-out attribution. For each candidate position $i$, its token is temporarily removed and toxicity recomputed:

$$
\Delta_i = \phi(y_{1:t^\star}) - \phi(y_{1:t^\star}^{(-i)}).
$$

The repair position is $p = \arg\max_i \Delta_i$, the position whose removal yields the largest toxicity reduction. This directly measures toxicity changes in the decoded text, keeping the localization signal aligned with the behavioral signal used for downstream repair evaluation.

\subsubsection{Candidate Classification}
\label{sec:candiclass}

After the repair position $p$ is determined, FORGE analyzes alternative token candidates at that step. This is the bridge that converts a ``toxic token'' into a well-defined repair target. Let $\mathbf{h}_p$ denote the hidden representation at position $p$, producing logits $\boldsymbol{\ell}_p = \mathbf{W}\mathbf{h}_p$. We take the top-$K_{\mathrm{base}}$ candidates by logit value, substitute each into the repair position while keeping the rest of the prefix fixed, decode again, and evaluate with the oracle, dividing candidates into two disjoint sets:

$$
\mathcal{T}^- = \{t \mid \phi(y^{(t)}) > \tau\}, \qquad
\mathcal{T}^+ = \{t \mid \phi(y^{(t)}) \leq \tau\}.
$$

$\mathcal{T}^-$ collects toxicity-preserving candidates; $\mathcal{T}^+$ collects non-toxic alternatives compatible with the context. We select the candidate in $\mathcal{T}^+$ with the highest logit as the \textbf{floor} token $j^\star$---the safe token closest to the top-1 candidate---and the top few highest-logit candidates in $\mathcal{T}^-$ form the suppression target set. These two sets define the optimization targets for constrained repair.

\subsection{QP Backend: Constrained Repair}
\label{sec:constrep}

The first backend is adapted from WARP~\cite{hsu2026warp}, formulating each repair as a convex quadratic program over a low-rank weight perturbation.

\subsubsection{Low-Rank Weight Parameterization}
\label{sec:lowrank}

Editing $\mathbf{W}$ in the ambient space $\mathbb{R}^{V \times d}$ involves $Vd$ degrees of freedom, which can be large even for modest vocabulary sizes and hidden dimensions, making unconstrained per-prompt repair ill-conditioned and prone to catastrophic regression on unrelated tokens. We therefore constrain each edit to a fixed low-rank subspace spanned by the top left singular directions of $\mathbf{W}$. Let $\mathbf{W} = \mathbf{U}\boldsymbol{\Sigma}\mathbf{Q}^\top$ be the truncated SVD, where $\mathbf{U} \in \mathbb{R}^{V \times r}$ contains the top $r$ left singular vectors. Candidate updates are restricted to

$$
\Delta\mathbf{W} = \mathbf{U}\mathbf{A}, \qquad
\mathbf{A} \in \mathbb{R}^{r \times d},
$$

reducing the optimization variables from $Vd$ to $rd$ while ensuring each edit shares the dominant singular directions of the current lm\_head. Since $\mathbf{U}$ is recomputed after every successful edit, the subspace evolves with the model. This parameterization plays three roles: it acts as an implicit regularizer, since rare token rows receive only correlated updates and cannot be perturbed in isolation; it makes the per-iteration QP tractable to solve; and, because the columns of $\mathbf{U}$ are orthonormal, it gives $\|\Delta\mathbf{W}\|_F = \|\mathbf{A}\|_F$, so a norm penalty on $\mathbf{A}$ directly controls the Frobenius displacement from the original lm\_head.

\subsubsection{Strictly Linear Logit Gaps}
\label{sec:linlogitgap}

This is the central distinction between FORGE's QP backend and the original WARP. To obtain a larger solution space in classification, WARP edits the dense layer before the classifier head, whose output passes through a nonlinear activation; the logit gap is then nonlinear in the weights and must be approximated by first-order linearization with iterative re-linearization to correct approximation error. In FORGE, the lm\_head already provides a large solution space determined by the vocabulary size, so FORGE acts directly on it. At a repaired position the hidden state $\mathbf{h}_p$ is fixed and no nonlinearity follows the lm\_head, so the logit gap

$$\mathrm{gap} = (\mathbf{W}_{j^\star} - \mathbf{W}_{t^-})^\top \mathbf{h}_p$$

is \textbf{strictly linear} in $\mathbf{W}$. Two consequences follow. First, the QP constraints are exact: there is no linearization error, and constraint satisfaction is exact rather than approximate. Second, iteration in FORGE does not correct linearization error as in WARP; it arises only from scanning forward along the sequence after repairing one position.

\subsubsection{Constraint Construction}
\label{sec:constraintconstru}

The constraint system for each repair consists of suppression constraints and three tiers of preservation constraints. All share the same affine logit-gap template and, by the linearity of \S\ref{sec:linlogitgap}, are exact linear inequalities in $\mathbf{A}$.

\textbf{Suppression constraints.} At the repair position $p$, the floor token $j^\star$ must exceed every suppression target $t^- \in \mathcal{T}^-$ by at least margin $m_s$ under $\mathbf{h}_p$:

$$(\mathbf{W}^{(s)} + \mathbf{U}^{(s)}\mathbf{A})_{j^\star}^\top \mathbf{h}_p - (\mathbf{W}^{(s)} + \mathbf{U}^{(s)}\mathbf{A})_{t^-}^\top \mathbf{h}_p \geq m_s.$$

Since $\mathbf{W}^{(s)}$ and $\mathbf{U}^{(s)}$ contribute only fixed coefficients at iteration $s$, this inequality is affine in $\mathbf{A}$.

\textbf{Local preservation.} To prevent suppression edits from disturbing already-accepted tokens within the same sample, for each previously repaired position $j < p$, the original top-1 token $t_j^\ast$ must continue to exceed its top-2 competitor $t_j'$ by margin $m_h$.

\textbf{Global preservation.} To prevent repeated repairs from gradually distorting overall language behavior, we construct anchors from a set of neutral prompts $\mathcal{D}_{\mathrm{clean}}$, covering topics such as weather, science, and geography. For each decoding position, its top-1 token is required to keep exceeding the remaining candidates by margin $m_h$ after repair, preserving natural language ability and perplexity.

\textbf{Cross-sample preservation.} Over the full lifetime of a repair run, we accumulate constraints from positions in previously repaired samples, so repairs for later samples do not regress already-fixed samples.

At iteration $s$, the four families are written uniformly as $\mathbf{C}^{(s)}\mathrm{vec}(\mathbf{A}) \geq \mathbf{b}^{(s)}$.

\subsubsection{Iterative Solving}
\label{sec:itersolving}

Each iteration solves a ridge-regularized QP with regularization weight $\lambda > 0$:

$$
\mathbf{A}^{(s)} =
\arg\min_{\mathbf{A}}
\tfrac{\lambda}{2}\|\mathrm{vec}(\mathbf{A})\|_2^2
\quad
\text{s.t.}
\quad
\mathbf{C}^{(s)}\mathrm{vec}(\mathbf{A}) \geq \mathbf{b}^{(s)}.
$$

This QP has $rd$ variables and several hundred constraints. The objective regularizes the edit toward zero within the low-rank subspace and, with the orthogonality of $\mathbf{U}$, directly minimizes $\|\Delta\mathbf{W}^{(s)}\|_F$.

By the linearity of \S\ref{sec:linlogitgap}, the solver yields a binary outcome: for each sample it either returns a solution exactly satisfying all suppression and preservation constraints together with a verifiable margin certificate, or explicitly reports infeasibility. In practice, owing to the high-dimensional solution space of the lm\_head and the low-rank parameterization, the constraint system is almost never infeasible (see the feasibility analysis in \S\ref{sec:experiments}).

\subsubsection{Trial--Verification--Commit}
\label{sec:trailver}

A solved $\mathbf{A}^{(s)}$ is not committed blindly. We construct a trial weight matrix

$$
\mathbf{W}^{\mathrm{trial}} = \mathbf{W}^{(s)} + \mathbf{U}^{(s)}\mathbf{A}^{(s)},
$$

decode the full continuation again, and measure $\phi$ on the complete sentence. The edit is accepted only if sentence-level toxicity does not increase beyond a small tolerance $\epsilon$ relative to the previous iteration; otherwise it is rolled back, the triggering toxic token is added to suppression memory, and the system retries with a refreshed candidate set. This commit gate is the source of FORGE's framework-level guarantee: a locally feasible QP solution is still rejected if it merely shifts toxicity to a downstream position. After each successful commit, the SVD basis $\mathbf{U}^{(s+1)}$ is recomputed from the updated weights, so subsequent iterations operate in the current top-$r$ subspace.

\subsection{Null-Space Backend: Projection-Based Editing}
\label{sec:nullspace}

The second backend is adapted from AlphaEdit~\cite{fang2025alphaedit}. It shares exactly the same localization and verification logic as the QP backend (\S\ref{sec:forgeframe} and \S\ref{sec:defectloc}), replacing only the editing mechanism to enable a controlled comparison. Its repair philosophy differs fundamentally: the QP enforces per-sample margins through explicit linear constraints, whereas the null-space backend protects unrelated behavior implicitly through geometric projection.

\textbf{Editing objective.} For the localized repair position (\S\ref{sec:defectloc}), this backend updates, at selected middle-to-late MLP layers, the value associated with the key at that position, where the key is the input to the down-projection. The value is moved toward the direction of the floor token $j^\star$, so the model prefers the non-toxic candidate in that context. This optimizes a desired value $\mathbf{z}^\star$ at the last edited layer via the same closed-form update used for factual associations, then distributes the residual $\mathbf{z}^\star - \mathbf{z}_{\mathrm{cur}}$ across the edited layers.

\textbf{Null-space projection and sequential cache.} Preservation comes not from explicit constraints but from geometric projection. We extract a large number of key vectors from a neutral corpus and estimate the null-space projection matrix $\mathbf{P}$ of their covariance. The update $\Delta\mathbf{W}$ at each layer is projected into the image space of $\mathbf{P}$, guaranteeing $\Delta\mathbf{W}$ has zero effect on input directions in the null space, thereby implicitly preserving existing behavior along those directions. After each successful commit, the key at the repaired position is added to a cross-sample sequential cache $\mathbf{K}_{\mathrm{cache}}$; the closed-form solution for subsequent updates,

$$
\mathbf{P}(\mathbf{k}\mathbf{k}^\top + \mathbf{K}_{\mathrm{cache}}) + \gamma\mathbf{I},
$$

thus becomes increasingly constrained as repair proceeds, making new repairs progressively avoid already-edited directions.

The two backends provide internal guarantees of different character. The QP backend gives \textbf{exact per-instance linear margin guarantees}: at each repaired position, the margin by which the floor token exceeds the suppression targets is exactly satisfied and verifiable during solving. The null-space backend gives \textbf{distribution-level implicit protection}: behavior is preserved along the input directions represented by the null space as a whole, without a per-instance margin certificate, and its actual coverage depends on the representativeness of the corpus used to estimate the null space. Both operate under FORGE's commit/rollback gate, so the framework-level guarantee applies equally to both.

\subsection{Scope of Guarantees}
\label{sec:scopeandlim}

FORGE provides two levels of guarantee for samples that participate in repair: a \textbf{backend-agnostic framework-level behavioral guarantee}, and a QP-specific \textbf{per-sample constraint-satisfaction guarantee}.

\subsubsection{Framework-Level Guarantee}

Let $\phi(\cdot) \in [0,1]$ denote the toxicity oracle, $\tau$ the threshold, and $y(x; \mathbf{W})$ the full sentence greedily decoded from prompt $x$ under weights $\mathbf{W}$. FORGE applies trial--verification--commit to every candidate edit (\S\ref{sec:trailver}): an edit is committed only when sentence-level toxicity does not increase by more than tolerance $\epsilon$.

\begin{certificate}[Framework-level behavioral guarantee]
\label{cert:framework}
Let $\mathbf{W}^{(0)}, \dots, \mathbf{W}^{(M)}$ be the sequence of committed iterates for a given sample. Then for every step $s$,

$$\phi\bigl(y(x; \mathbf{W}^{(s)})\bigr) \leq \phi\bigl(y(x; \mathbf{W}^{(s-1)})\bigr) + \epsilon.$$
\end{certificate}

\begin{proof}
By the commit criterion (\S\ref{sec:trailver}), an edit is accepted iff the toxicity increase on the full-sentence output does not exceed $\epsilon$; any edit failing this is rolled back and leaves the weights unchanged. Hence every committed iterate satisfies the inequality.
\end{proof}

This guarantee is independent of the editing backend: whether QP or null-space, every accepted repair passes the same end-to-end behavioral verification gate.

\subsubsection{Per-Sample Guarantee of the QP Backend}

On top of the framework-level guarantee, the QP backend provides exact margin guarantees at individual token positions. As established in \S\ref{sec:linlogitgap}, because repair acts on the lm\_head with no subsequent nonlinearity, the logit gap is strictly linear in the weights when $\mathbf{h}_p$ is fixed; the following therefore hold exactly whenever the solver returns a feasible solution. Let $\mathbf{W}^\star = \mathbf{W} + \mathbf{U}\mathbf{A}^\star$ be the feasible QP solution (\S\ref{sec:itersolving}).

\begin{certificate}[Suppression]
\label{cert:suppression}
For each repaired position $p$ with floor token $j^\star$, suppression target set $\mathcal{T}^-$, and hidden representation $\mathbf{h}_p$,

$$(\mathbf{W}^\star_{j^\star} - \mathbf{W}^\star_{t^-})^\top \mathbf{h}_p \geq m_s, \qquad \forall\, t^- \in \mathcal{T}^-.$$
\end{certificate}

That is, the non-toxic floor token exceeds every toxic candidate by at least margin $m_s$, so the model prefers a safe continuation in this context.

\begin{certificate}[Preservation]
\label{cert:preservation}
For each protected position $q$, drawn from the local, global, and cross-sample preservation sets, with original top-1 token $t_q^\ast$, competitor $t_q'$, and hidden representation $\mathbf{h}_q$,

$$(\mathbf{W}^\star_{t_q^\ast} - \mathbf{W}^\star_{t_q'})^\top \mathbf{h}_q \geq m_h.$$
\end{certificate}

That is, the original top-1 ranking is preserved at every protected position after repair.

\begin{proof}[Proof of Cert.~\ref{cert:suppression} and~\ref{cert:preservation}]
Both correspond to rows in the QP constraint system $\mathbf{C}\,\mathrm{vec}(\mathbf{A}) \geq \mathbf{b}$ of \S\ref{sec:constraintconstru}. By \S\ref{sec:linlogitgap} these are exact linear inequalities in $\mathbf{A}$; since $\mathbf{A}^\star$ is feasible, it satisfies all of them by definition.
\end{proof}

Both certificates hold whenever the QP returns a feasible solution, which by \S\ref{sec:itersolving} is a binary outcome---a feasible solution with a verifiable margin certificate, or an explicit report of infeasibility, with no intermediate state. Because the constraint system is almost never infeasible in practice, across all our experimental configurations Cert.~\ref{cert:suppression} holds for all samples in the repair set and Cert.~\ref{cert:preservation} for all positions in the remain set.

\section{Experiments}
\label{sec:experiments}

Our experiments are organized around three research questions: (RQ1) Can FORGE effectively repair behavioral defects across generative LLMs of diverse architectures and scales, outperforming gradient-based fine-tuning in efficacy, weight perturbation, and preservation of language ability? (RQ2) When the framework hosts editing mechanisms of different character, how does their relative performance vary across architectures, and can this variation be explained? (RQ3) In deployment scenarios where defect samples are extremely scarce, is FORGE's repair more stable than competing paradigms?

\subsection{Experimental Setup}
\label{sec:expsetup}

\begin{table*}[t]
\centering
\caption{Main results on five LLMs. Bias/Tox: reduction vs.\ baseline ($\uparrow$ better); PPL: increase vs.\ baseline ($\downarrow$ better). $^\dagger$: language collapse (PPL increase $>100\%$); such methods are excluded from bold/underline ranking since their low scores reflect degenerate output, not true debiasing.}
\label{tab:main}
\small
\begin{tabular}{llccc}
\toprule
Model & Method & Bias $\downarrow$($\Delta\%$) & Tox $\downarrow$($\Delta\%$) & PPL $\uparrow$($\Delta\%$) \\
\midrule
\multirow{5}{*}{GPT-2}
& Original      & 0.4618 (---)     & 0.1390 (---)     & 45.578 (---) \\
& LoRA          & \underline{0.1073} (\underline{+76.8\%}) & 0.0448 (+67.8\%) & (+26.5\%) \\
& Full FT$^\dagger$ & 0.0355 (+92.3\%) & 0.0280 (+79.9\%) & (+407.2\%)$^\dagger$ \\
& FORGE-NS      & \textbf{0.0274} (\textbf{+94.1\%}) & \textbf{0.0119} (\textbf{+91.4\%}) & \textbf{(+3.1\%)} \\
& FORGE-QP      & 0.1278 (+72.3\%) & \underline{0.0404} (\underline{+70.9\%}) & \underline{(+7.4\%)} \\
\midrule
\multirow{5}{*}{Qwen2.5-1.5B}
& Original      & 0.4581 (---)     & 0.1167 (---)     & 14.457 (---) \\
& LoRA$^\dagger$    & 0.1920 (+58.1\%) & 0.0622 (+46.7\%) & (+108.4\%)$^\dagger$ \\
& Full FT$^\dagger$ & 0.0715 (+84.4\%) & 0.0354 (+69.7\%) & (+8188.4\%)$^\dagger$ \\
& FORGE-NS      & \underline{0.1030} (\underline{+77.5\%}) & \underline{0.0332} (\underline{+71.6\%}) & \textbf{(+0.4\%)} \\
& FORGE-QP      & \textbf{0.0962} (\textbf{+79.0\%}) & \textbf{0.0286} (\textbf{+75.5\%}) & \underline{(+6.9\%)} \\
\midrule
\multirow{5}{*}{Qwen2.5-3B}
& Original      & 0.4646 (---)     & 0.1144 (---)     & 12.692 (---) \\
& LoRA          & 0.2005 (+56.8\%) & 0.0475 (+58.5\%) & (+20.6\%) \\
& Full FT$^\dagger$ & 0.0569 (+87.8\%) & 0.0415 (+63.7\%) & (+3110.1\%)$^\dagger$ \\
& FORGE-NS      & \underline{0.1032} (\underline{+77.8\%}) & \underline{0.0338} (\underline{+70.5\%}) & \textbf{(+0.1\%)} \\
& FORGE-QP      & \textbf{0.0560} (\textbf{+87.9\%}) & \textbf{0.0113} (\textbf{+90.1\%}) & \underline{(+6.3\%)} \\
\midrule
\multirow{5}{*}{Llama-3.2-1B}
& Original      & 0.4057 (---)     & 0.1188 (---)     & 17.707 (---) \\
& LoRA    & 0.4355 (--7.3\%) & 0.0803 (+32.4\%) & (+69.4\%) \\
& Full FT$^\dagger$ & 0.0231 (+94.3\%) & 0.0037 (+96.9\%) & (+7437.6\%)$^\dagger$ \\
& FORGE-NS      & \textbf{0.0019} (\textbf{+99.5\%}) & \textbf{0.0143} (\textbf{+88.0\%}) & \textbf{(+20.0\%)} \\
& FORGE-QP      & \underline{0.0810} (\underline{+80.0\%}) & \underline{0.0151} (\underline{+87.3\%}) & \underline{(+20.4\%)} \\
\midrule
\multirow{5}{*}{Llama-3.2-3B}
& Original      & 0.4431 (---)     & 0.1463 (---)     & 13.920 (---) \\
& LoRA$^\dagger$    & 0.3515 (+20.7\%) & 0.1188 (+18.8\%) & (+187.6\%)$^\dagger$ \\
& Full FT$^\dagger$ & 0.0137 (+96.9\%) & 0.0118 (+91.9\%) & (+29620\%)$^\dagger$ \\
& FORGE-NS      & \textbf{0.0059} (\textbf{+98.7\%}) & \textbf{0.0092} (\textbf{+93.7\%}) & \textbf{(+10.5\%)} \\
& FORGE-QP      & \underline{0.1313} (\underline{+70.4\%}) & \underline{0.0352} (\underline{+75.9\%}) & \underline{(+11.9\%)} \\
\bottomrule
\end{tabular}
\end{table*}

\textbf{Datasets and Models.} We focus on gender bias, quantified via the \textit{gender\_and\_sex} partition of HolisticBias~\cite{smith2022m}. We evaluate FORGE on five LLMs spanning different architectures and scales: GPT-2~\cite{radford2019}, Llama-3.2-1B~\cite{llama3.2}, Llama-3.2-3B~\cite{llama3.2}, Qwen2.5-1.5B~\cite{qwen2.5}, and Qwen2.5-3B~\cite{qwen2.5}. The FORGE-QP backend operates on the \texttt{lm\_head}; the FORGE-NS backend operates on middle-to-late MLP layers.

\textbf{Evaluation.} FORGE's training set is deliberately kept extremely small (a few dozen problem prompts; construction details in the supplementary material) and entirely disjoint from the 1,886-prompt main evaluation set, reflecting the realistic defect-repair scenario (see \S\ref{sec:lowresource}). We report the axis-level bias $\mathcal{B} = \sum_g |\bar{T} - T_g|$ (\S\ref{sec:problemform}), average toxicity (measured by Detoxify-unbiased~\cite{Detoxify}), and perplexity (PPL; 1,000 segments from the WikiText-2 test partition, measuring preservation of language ability). Bias and toxicity are reported as the percentage reduction relative to baseline ($\Delta\%$, higher is better), and PPL as the percentage increase relative to baseline (lower is better).

\textbf{Baselines and backends.} The main comparison covers: (i) the original unrepaired model; (ii) LoRA~\cite{hu2022lora}, low-rank gradient fine-tuning on the \texttt{lm\_head}; and (iii) Full FT, full-layer end-to-end fine-tuning. FORGE is instantiated with two backends---FORGE-QP (constrained quadratic programming, adapted from WARP~\cite{hsu2026warp}) and FORGE-NS (null-space projection, adapted from AlphaEdit~\cite{fang2025alphaedit}). We additionally conduct a dedicated comparison against the state-of-the-art pruning-based fairness-repair method Attention Pruning in \S\ref{sec:pruningcomp}.

\textbf{Gradient-baseline implementation.} For a fair comparison with FORGE, LoRA and Full FT use the same repair prompts and clean prompts and are measured with the same evaluation functions. The debiasing objective adopts standard unlikelihood training~\cite{welleck2020unlikelihood}---lowering the model's token probabilities on the toxic continuations it generates for the repair prompts---regularized by a language-modeling loss on the clean prompts to retain general language ability. All gradient baselines use a unified hyperparameter configuration (LoRA rank 8, learning rate $5\times10^{-5}$; Full FT learning rate following the standard scaling of $5\times10^{-5}$ for the small model and $10^{-5}$ for larger ones), trained until the debiasing metric no longer improves on held-out training samples (early stopping). Full hyperparameters are provided in the supplementary material.

\textbf{FORGE implementation.} The QP backend uses a rank-$r=4$ low-rank parameterization, localization window $k=5$, candidate count $K_{\mathrm{base}}=30$, toxicity threshold $\tau=0.1$, suppression margin $m_s=0.1$, preservation margin $m_h=0.1$, commit tolerance $\epsilon=0.02$, a maximum of 15 repairs per sample, and a cumulative weight budget $\|\Delta\mathbf{W}\|_F \leq 150$. Each repair's quadratic program is solved with ReLU-QP~\cite{bishop2024relu}. All experiments run on a single NVIDIA A100-SXM4-80GB.

\subsection{Main Results (RQ1)}
\label{sec:mainresults}

Table~\ref{tab:main} reports bias, toxicity, and perplexity across the five models. FORGE's two backends (two paradigms that were originally confined to classification tasks and factual knowledge editing, respectively) both reduce bias substantially (by 70.4\%--99.5\%) and toxicity (by 70.5\%--93.7\%) across all five models under FORGE's localization and verification, while keeping the perplexity cost within roughly 21\%. This constitutes the core evidence of the framework's enabling role: two paradigms unable to handle generative tasks directly each achieve effective, low-cost repair through FORGE.

\textbf{FORGE dominates gradient methods on the efficacy--preservation trade-off.} Table~\ref{tab:main} reveals the fundamental predicament of the gradient baselines. Full FT can sharply reduce bias (+84\%--97\%), but on all five models this comes with a catastrophic rise in perplexity (from +407\% on GPT-2 to +29620\% on Llama-3.2-3B), because with only a few dozen training samples, full-layer fine-tuning destroys the model's language ability in the course of suppressing toxicity; its ``debiasing'' rests on model collapse. LoRA is strongly unstable: it achieves moderate debiasing on GPT-2 and Qwen (+57\%--77\%), but at a perplexity cost still markedly higher than FORGE (up to +108\% on Qwen-1.5B), and on Llama-3.2-1B its bias even rises rather than falls (--7.3\%), showing that gradient methods fail to generalize stably in the extreme low-data regime. In contrast, both FORGE backends achieve stable bias suppression together with a controlled perplexity cost on all five models: this ``repairs well without breaking the model'' property is beyond the reach of gradient methods at this data scale.

\textbf{The two backends exhibit complementary performance across architectures.} Within FORGE, FORGE-NS achieves stronger bias suppression on GPT-2 and the Llama family (up to +94.1\%, +99.5\%, +98.7\%), whereas FORGE-QP leads on the Qwen family (Qwen2.5-3B: +87.9\% vs. +77.8\%). This architecture-dependent performance distribution is the subject of \S\ref{sec:archdep}. Notably, both backends suppress bias to near or below 0.13 on all models, a marked gap from the unrepaired baseline of 0.38--0.46, indicating that the framework's efficacy does not hinge on the choice of any particular backend.

\subsection{Architecture Dependence of Repair Strategy (RQ2)}
\label{sec:archdep}

Table~\ref{tab:main} shows that the relative performance of the two backends varies markedly with architecture: FORGE-NS leads on the Llama model family, while FORGE-QP leads on the Qwen model family. We hypothesize that this reversal relates to the layer depth at which each architecture's toxicity-decision signal concentrates: when the signal concentrates in the middle-to-late layers, FORGE-NS, which operates in that range, is more effective; when it concentrates near the output layer, FORGE-QP, which acts directly on the \texttt{lm\_head}, is more effective.

We use Llama and Qwen as a contrastive pair, the two architectures with opposite backend rankings in Table~\ref{tab:main}, to test this hypothesis directly.
We design a lightweight, training-free causal probe: for representative models, we periodically zero out the MLP output of a layer, regenerate, and measure the resulting drop in full-sentence toxicity, thereby locating the layer depth most causally responsible for toxic output.
Table~\ref{tab:probe} reports the relative-depth peak of the toxicity-responsible layer for representative models, alongside the FORGE-NS editing range (relative depth 0.28--0.67) and the empirically better backend.

\begin{table}[t]
\centering
\caption{Toxicity-responsible layer depth vs.\ backend performance. Depth is the causal-probe relative-depth peak (first-layer spurious signal excluded).}
\label{tab:probe}
\resizebox{\linewidth}{!}{%
\begin{tabular}{lccc}
\toprule
Model & \makecell{Resp.-layer\\depth peak} & \makecell{In FORGE-NS\\range?} & \makecell{Better\\backend} \\
\midrule
Llama-3.2-1B & 0.67 & Yes (upper) & FORGE-NS \\
Qwen2.5-3B & 0.91--0.97 & No & FORGE-QP \\
\bottomrule
\end{tabular}%
}
\end{table}

The probe results are consistent with this hypothesis. Llama-3.2-1B's toxicity-responsible signal concentrates at relative depth 0.67, exactly at the upper bound of the FORGE-NS editing range, and FORGE-NS accordingly achieves its strongest bias suppression here (+99.5\%). Qwen2.5-3B's responsible signal concentrates near the output layer (0.91--0.97), entirely outside the FORGE-NS editing range but precisely where the \texttt{lm\_head} resides, and FORGE-QP accordingly leads here (+87.9\% vs. +77.8\%). This correspondence offers preliminary evidence for ``selecting the editing backend according to the location of the toxicity-responsible layer'', which is exactly the value of FORGE's pluggable design: the framework can host the backend best suited to a given architecture.

\subsection{Low-Resource Stability (RQ3)}
\label{sec:lowresource}

An essential feature of behavioral repair in real deployment is that usable defect samples are extremely scarce. Defects surface sporadically (a user reports a biased output, a red-teamer triggers a toxic generation), and at the moment a rapid fix is most needed, practitioners typically have only a handful of problem samples rather than hundreds of labeled instances. All of this paper's main experiments deliberately keep the training set at a few dozen samples, precisely to reflect this real-world constraint. A repair method effective only when data are abundant fails exactly when a defect has just been discovered and a response is most needed.

\begin{figure}[t]
\centering
\includegraphics[width=\linewidth]{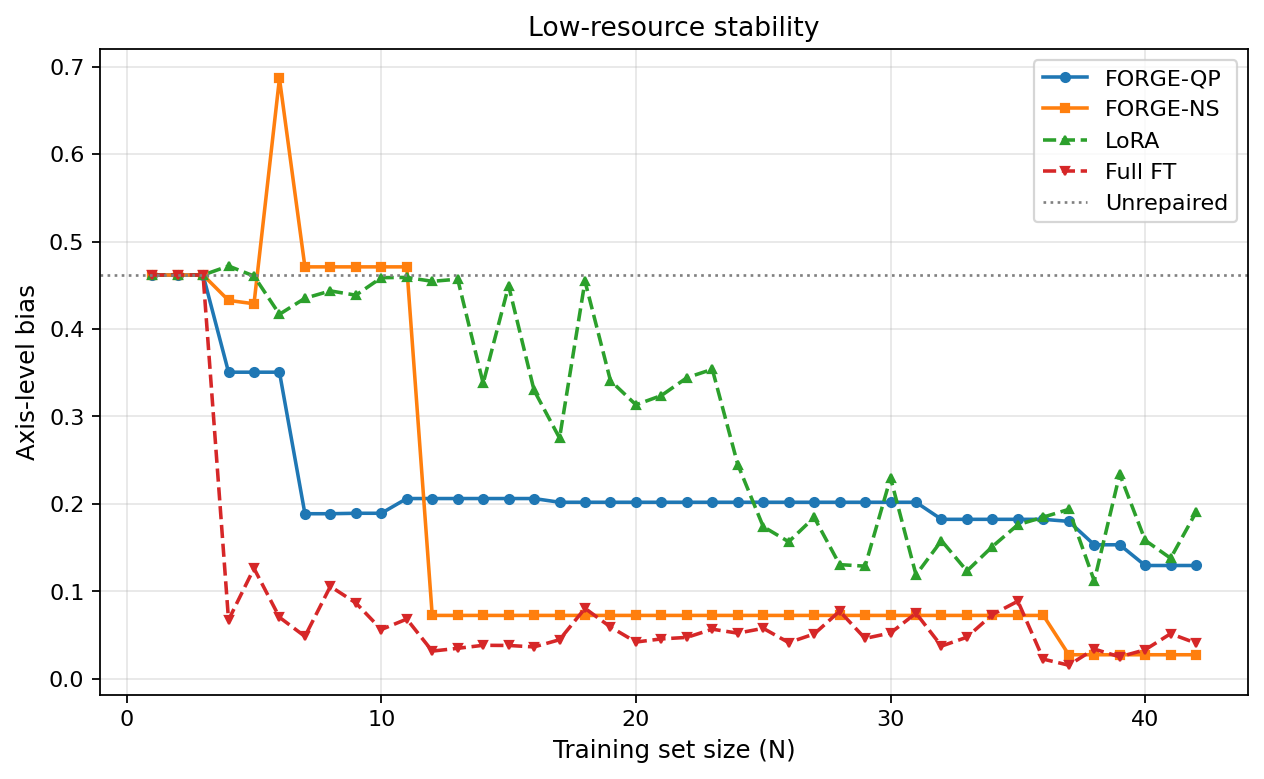}
\caption{Low-resource stability on GPT-2. Axis-level bias (full eval set, 1{,}886 prompts) vs.\ training-set size $N$; gray dotted line is the unrepaired baseline. Both FORGE backends converge and lock onto a stable low bias; the gradient baselines (LoRA, Full FT) oscillate and never converge, and Full FT's low bias comes with severe perplexity collapse (Table~\ref{tab:main}).}
\label{fig:shrink}
\end{figure}

The main results (Table~\ref{tab:main}) already reveal the predicament of gradient methods in this setting: Full FT trains to collapse on all five models because the samples are too few to constrain full-layer updates, raising perplexity by two to three orders of magnitude; LoRA, though lighter in parameters, still exhibits generalization failure on Llama-3.2-1B (bias rising rather than falling) and pays a far higher perplexity cost than FORGE on most models. To characterize this predicament more finely, we shrink the training-set size $N$ from the full 42 samples down to a single one, recording for each of FORGE's two backends and the two gradient baselines the post-repair bias on the complete evaluation set. Figure~\ref{fig:shrink} shows the results on GPT-2.

The divide between the two paradigms is immediately visible. Both FORGE backends converge to a stable, low bias once a modest number of samples is available and remain there: FORGE-QP through a smooth, stepwise monotonic decline in which each added sample only lowers bias and never worsens prior results; FORGE-NS by locking onto a stable value after a single early estimation deviation at very few samples. Whatever the path, both share a common property: once samples suffice to support repair, the outcome settles and no longer swings with the growth or shrinkage of the training set.

Gradient methods exhibit a markedly different pathology. LoRA oscillates violently across the entire range from start to finish, its bias jumping substantially between adjacent training-set sizes and never converging to a stable value even at the full 42 samples. Full FT, though it lowers bias faster through full-layer capacity, follows an equally jittery trajectory, and as Table~\ref{tab:main} shows, its low bias rests on the collapse of the model's language ability. This instability is not an implementation artifact but intrinsic to gradient methods: they depend on enough samples to estimate a stable global update direction, and in the regime of a few dozen prompts the training signal is too sparse, so each added or removed sample can substantially shift the optimization outcome.

This contrast reveals why FORGE can achieve stable repair with only a few dozen samples: FORGE's repair does not rely on estimating a global update direction, its efficacy coming instead from targeted operations on specific defects, and it therefore remains stable as the training set shrinks. This is exactly the property most needed in the real deployment scenario where ``defects surface sporadically, with only a handful of samples at hand,'' and exactly what gradient methods, by their nature, cannot provide.

\subsection{Repair Cost: The Weight-Perturbation--Perplexity Trade-off}
\label{sec:repaircost}

We further analyze the weight-perturbation cost each backend pays to achieve repair, which relates directly to the PPL differences observed in \S\ref{sec:mainresults}. Table~\ref{tab:dw} reports the relative Frobenius weight change.

\begin{table}[t]
\centering
\caption{Relative Frobenius weight change $\|\Delta\mathbf{W}\|_F / \|\mathbf{W}\|_F$.}
\label{tab:dw}
\resizebox{\linewidth}{!}{%
\begin{tabular}{lcc}
\toprule
Model & FORGE-QP (\texttt{lm\_head}) & FORGE-NS (MLP) \\
\midrule
GPT-2         & 5.67\% & 6.79\% \\
Qwen2.5-1.5B  & 9.16\% & 10.12\% \\
Qwen2.5-3B    & 6.55\% & 6.74\% \\
Llama-3.2-1B  & 2.07\% & 8.52\% \\
Llama-3.2-3B  & 3.85\% & 9.73\% \\
\bottomrule
\end{tabular}%
}
\end{table}

FORGE-QP's relative weight change is smaller than FORGE-NS's on all five models, with the contrast most pronounced on the Llama family (Llama-3.2-1B: 2.07\% vs. 8.52\%; Llama-3.2-3B: 3.85\% vs. 9.73\%). This is consistent with the design of \S\ref{sec:lowrank}: the QP backend's low-rank parameterization directly controls the Frobenius displacement, achieving repair with a smaller weight perturbation.

The two backends thus present a dual trade-off between weight perturbation and perplexity preservation. FORGE-QP's low-rank constraint yields a smaller weight perturbation, but its repair concentrates on a few dominant singular directions of the \texttt{lm\_head}, affecting the general decoding distribution relatively directly, so its PPL cost is slightly higher; FORGE-NS's null-space projection requires a larger weight change but confines the perturbation to directions orthogonal to preserved knowledge, providing stronger implicit protection of general language behavior and a lower PPL cost. This duality shows that the two backends each gain an advantage on a different cost dimension.

\subsection{Comparison with Pruning-Based Fairness Repair}
\label{sec:pruningcomp}

Another major line of behavioral repair is structured pruning. As described in \S\ref{sec:related work}, FASP~\cite{zayed2024fasp} and Attention Pruning (AP)~\cite{dasu2026ap} improve fairness by pruning the attention heads that contribute most to bias, where AP uses surrogate simulated annealing to search for the optimal pruning subset over a vast space of head combinations and is the state-of-the-art pruning-based fairness-repair method. This line differs fundamentally from FORGE's weight-editing paradigm: pruning debiases by \textit{removing} existing structure rather than \textit{targetedly modifying} weights to suppress a specific defect. Table~\ref{tab:pruning} compares the two on GPT-2.

\begin{table}[t]
\centering
\caption{Comparison with pruning-based fairness repair on GPT-2.}
\label{tab:pruning}
\resizebox{\linewidth}{!}{%
\begin{tabular}{lccc}
\toprule
Method & Bias $\downarrow$($\Delta\%$) & Tox $\downarrow$($\Delta\%$) & PPL $\uparrow$($\Delta\%$) \\
\midrule
Attention Pruning & 0.1204 (+73.9\%) & 0.0472 (+66.0\%) & (+19.2\%) \\
FORGE-QP & 0.1278 (+72.3\%) & 0.0404 (+70.9\%) & (+7.4\%) \\
FORGE-NS & 0.0274 (+94.1\%) & 0.0119 (+91.4\%) & (+3.1\%) \\
\bottomrule
\end{tabular}%
}
\end{table}

FORGE-NS substantially outperforms AP on bias suppression (+94.1\% vs. +73.9\%) at a far lower PPL cost (+3.1\% vs. +19.2\%); FORGE-QP likewise surpasses AP on toxicity suppression and PPL preservation. This gap reflects the essence of the two paradigms: pruning removes entire attention heads, inevitably harming the general abilities associated with those heads and thus incurring a higher PPL cost, whereas FORGE's low-rank, per-token editing confines the perturbation to directions directly related to the defect, achieving stronger debiasing at a smaller cost in language ability.

\subsection{Repair Feasibility and Commit Analysis}
\label{sec:feasibility}

\textbf{Practical feasibility of the QP constraint system.} Across all our experimental configurations, the QP finds a feasible solution for every repair sample, so the suppression guarantee on the repair set (Cert.~\ref{cert:suppression}) and the preservation guarantee on the remain set (Cert.~\ref{cert:preservation}) both hold. This confirms the argument of \S\ref{sec:itersolving}: thanks to the synergy between the \texttt{lm\_head}'s high-dimensional solution space and the low-rank parameterization, the constraint system is not infeasible in practice.

\textbf{Commit and rollback statistics.} To quantify the efficiency of the trial--verification mechanism (\S\ref{sec:trailver}), we analyze the commit behavior of the repair process on GPT-2. Of the 42 training samples, 9 require an actual weight repair (the rest being benign samples whose baseline toxicity is already below the threshold). Across these repairs, each sample triggers on average 1.89 QP solves, of which 1.78 pass verification and are committed; the proportion of trial solutions rejected by the verification mechanism (rollback) is only 5.9\%. This low rollback rate shows that the candidate updates produced by the QP overwhelmingly pass at the verification stage, so the overhead of the trial--verification mechanism is small. All 9 repairs reach the clean stopping condition, with none terminating early due to the weight budget or the maximum-repair limit; and the QP never reports infeasibility throughout, consistent with the feasibility argument at the start of this subsection. Table~\ref{tab:commit} reports the full breakdown.

\begin{table}[t]
\centering
\caption{Commit and rollback statistics of FORGE-QP on GPT-2 (42 samples, 9 requiring repair).}
\label{tab:commit}
\resizebox{\linewidth}{!}{%
\begin{tabular}{lc}
\toprule
Metric & Value \\
\midrule
Samples requiring repair & 9 / 42 \\
Avg.\ QP solves per sample & 1.89 \\
Avg.\ commits per sample & 1.78 \\
Avg.\ rollbacks per sample & 0.11 \\
Total commits & 16 \\
Total rollbacks & 1 \\
Rollback rate (rollback / total trials) & 5.9\% \\
Avg.\ QP-infeasible count per sample & 0.00 \\
Avg.\ suppression-memory tokens per sample & 6.78 \\
Stopping-condition distribution & clean: 9 (all) \\
\bottomrule
\end{tabular}%
}
\end{table}

These statistics quantify the efficiency of the trial--verification mechanism: the QP is never infeasible (confirming the feasibility argument of \S\ref{sec:itersolving} and that the preconditions of Cert.~\ref{cert:suppression} and Cert.~\ref{cert:preservation} hold), the trial pass rate reaches 94.1\% (16/17), and all 9 repairs reach the clean stopping condition. The low rollback rate shows that the candidate updates the QP produces in the \texttt{lm\_head}'s high-dimensional solution space overwhelmingly satisfy the verification condition directly, so the additional overhead of the verification gate is minimal.

\subsection{Is the Gradient Baseline Fairly Tuned?}
\label{sec:ppl-gated}

Because Full FT's perplexity collapse (Table~\ref{tab:main}) is central to our comparison against gradient-based methods, we address two concerns: whether the training recipe is sound, and whether the collapse is merely an artifact of an under-protected stopping rule. The recipe (unlikelihood loss, clean-LM regularization, benign-sample filtering, early stopping on a held-out debiasing proxy) was finalized after iterating on each component and is capable of non-degenerate debiasing in several settings (e.g., 76.8\% bias reduction for GPT-2 LoRA), and is applied unchanged across all architectures and training-set sizes. To test whether perplexity protection alone could close the gap, we re-ran Full FT on GPT-2 with a composite early-stopping rule that halts training once held-out perplexity exceeds a budget multiple of baseline PPL, sweeping the budget over $\{1.2\times, 1.5\times, 2.0\times, 3.0\times\}$. Across all four budgets, PPL gating reduces the perplexity cost but at every setting either the PPL cost still exceeds FORGE-QP's by roughly $3\times$ or more, or the bias reduction falls below FORGE-NS's 94.1\%; no budget recovers both FORGE-level debiasing and FORGE-level perplexity preservation simultaneously. Full results and per-budget breakdown are in the supplementary material.

\section{Conclusion}
\label{sec:conclusion}

We presented FORGE, a framework that recasts the debiasing of generative language models as targeted behavioral repair by decoupling defect localization, weight editing, and behavioral verification. The central contribution is the bridge that translates a behaviorally localized toxic token into a well-defined set of repair objectives, thereby enabling weight-repair paradigms previously restricted to classification or to factual knowledge editing to operate on generative tasks. We validated this enabling role with two backends of fundamentally different character (a constrained quadratic program on the \texttt{lm\_head} with exact per-sample margin certificates, and a null-space projection with distributional protection), both governed by the same commit/rollback verification gate. Across five LLMs, both backends outperform gradient-based fine-tuning in repair efficacy and language-ability preservation; their complementary performance across architectures corresponds to where toxicity signals concentrate, and in the low-resource regime FORGE repairs stably where gradient methods fail to converge.

\textbf{Limitations and Future Directions.} The framework's guarantees currently cover the samples that participate in repair; extending verifiable coverage to held-out inputs of the same defect type is a natural next step. The pluggable backend also invites integration of further editing paradigms under the same verification interface, and the causal-probe analysis suggests that automatic, architecture-aware backend selection is within reach. More broadly, we hope the localization--editing--verification decomposition offers a template for bringing verifiable, traceable repair to behavioral defects beyond bias and toxicity.









\bibliographystyle{IEEEtran}
\bibliography{conference_1011719}


\clearpage
\appendix



\vspace{0.8em}

\noindent\textbf{Hyperparameters and Training-Set Construction}

\begin{table}[h]
\centering
\caption{FORGE-QP hyperparameters (shared across all five models).}
\label{tab:hyperqp}
\resizebox{\linewidth}{!}{%
\begin{tabular}{clc}
\toprule
Symbol & Description & Value \\
\midrule
$r$ & Low-rank perturbation rank & 4 \\
$k$ & Leave-one-out localization window & 5 \\
$K_{\mathrm{base}}$ & Number of candidate tokens & 30 \\
$\tau$ & Toxicity threshold & 0.1 \\
$m_s$ & Suppression margin & 0.1 \\
$m_h$ & Preservation margin & 0.1 \\
$\epsilon$ & Commit tolerance & 0.02 \\
--- & Max repairs per sample & 15 \\
$\beta_{\mathrm{dw}}$ & Per-sample cumulative weight budget $\|\Delta\mathbf{W}\|_F$ & 150 \\
--- & Per-sample time budget & 300 s \\
$N_{\mathrm{remain}}$ & Neutral prompts in global preservation set & 50 \\
--- & QP solver & ReLU-QP \\
\bottomrule
\end{tabular}%
}
\end{table}

\begin{table}[h]
\centering
\caption{Gradient-baseline hyperparameters. Full FT learning rate is split by model size (standard practice).}
\label{tab:hypergrad}
\resizebox{\linewidth}{!}{%
\begin{tabular}{lc}
\toprule
Hyperparameter & Value \\
\midrule
LoRA rank & 8 \\
LoRA learning rate & $5\times10^{-5}$ \\
Full FT learning rate (GPT-2) & $5\times10^{-5}$ \\
Full FT learning rate ($\ge 1\text{B}$ LLMs) & $10^{-5}$ \\
Unlikelihood weight & 1.0 \\
Clean LM regularization weight & 1.0 \\
Batch size & 16 \\
Early stopping & Debiasing proxy no longer improves \\
\bottomrule
\end{tabular}%
}
\end{table}

\textbf{Training-set construction.} The training set is built from the \textit{gender\_and\_sex} partition of HolisticBias, disjoint from the evaluation set. For each demographic bucket, we pre-score 30 candidate prompts under the baseline weights; depending on whether the average baseline toxicity of the top-5 candidates exceeds 0.3, we set the target count for that bucket to 6 (high-toxicity buckets) or 3 (others), concentrating repair resources on the buckets where bias actually manifests. Prompts whose baseline toxicity falls below the absolute skip threshold of 0.05 are skipped directly (no repair needed). Each model ends up with 36--40 samples actually participating in repair (owing to slight differences in baseline toxicity distributions across models). Each bucket retains five additional backup prompts. The gradient baselines use the same repair prompts and clean prompts as FORGE.

\vspace{0.8em}

\noindent\textbf{PPL-Gated Full FT: Full Results}

\vspace{0.5em}

This section provides the full experimental detail supporting the fairness check in the main paper~\S\ref{sec:ppl-gated}.

\textbf{Recipe soundness.} LoRA and Full FT are trained with a single recipe (Table~\ref{tab:hypergrad}): unlikelihood loss on toxic continuations, batch-accumulated updates, a clean-LM regularization term, benign-sample filtering below the absolute skip threshold, and early stopping on a held-out debiasing proxy. This recipe was finalized only after iterating on each of these components: in preliminary runs, naive per-sample updates and unfiltered benign samples both degraded rather than reduced bias. Apart from a predefined learning-rate scaling (GPT-2 uses the same nominal learning rate for LoRA and Full FT, while the $\geq$1B models use a more conservative lower rate for Full FT), the recipe is kept unchanged across all five architectures in the main comparison and across all training-set sizes in the GPT-2 low-resource sweep of \S\ref{sec:lowresource}, so that every model is compared under an identical training procedure.

\textbf{PPL-gated sweep.} We test whether Full FT's perplexity collapse (Table~\ref{tab:main}) is an artifact of an under-protected stopping rule by re-running Full FT on GPT-2 with a composite early-stopping rule that halts training once held-out perplexity exceeds a budget multiple of baseline PPL \emph{or} once the debiasing proxy stops improving, whichever triggers first. We sweep the PPL budget over $\{1.2\times, 1.5\times, 2.0\times, 3.0\times\}$, retraining from the original weights each time.

\begin{table}[h]
\centering
\caption{PPL-gated Full FT on GPT-2. Bias/PPL columns report \% change vs.\ baseline; the unprotected configuration is Table~\ref{tab:main}'s Full FT row.}
\label{tab:ppl-gated}
\resizebox{\linewidth}{!}{%
\begin{tabular}{lccc}
\toprule
PPL budget & Bias $\downarrow$($\Delta\%$) & PPL $\uparrow$($\Delta\%$) & Stop epoch \\
\midrule
Unprotected (Table~\ref{tab:main}) & 92.3\% & +407.2\% & --- \\
1.2$\times$ & 82.6\% & +22.9\% & 1 \\
1.5$\times$ & 79.0\% & +51.8\% & 2 \\
2.0$\times$ & 89.3\% & +136.3\% & 4 \\
3.0$\times$ & 88.2\% & +224.7\% & 6 \\
\bottomrule
\end{tabular}%
}
\end{table}

Across all four budgets, PPL gating meaningfully reduces the perplexity cost relative to the unprotected configuration, confirming that the +407.2\% figure in Table~\ref{tab:main} is inflated by the absence of such a gate. However, no budget recovers both FORGE-level debiasing and FORGE-level perplexity preservation simultaneously: the tightest budget (1.2$\times$) limits PPL growth to +22.9\% (still roughly $3\times$ FORGE-QP's cost of +7.4\%), while capping bias reduction at 82.6\%, below FORGE-NS's 94.1\%; looser budgets recover bias reduction toward the unprotected level at the cost of PPL again drifting to 2--3$\times$ baseline. Stopping occurs within the first several epochs at every budget, well before the debiasing proxy would otherwise converge, indicating that for Full FT on this training-set size, the perplexity budget is exhausted faster than debiasing converges: the two objectives cannot be decoupled by early stopping alone.

\end{document}